\documentclass[conference]{IEEEtran}
\IEEEoverridecommandlockouts

\usepackage{cite}
\usepackage{amsmath,amssymb,amsfonts}
\usepackage{algorithmic}
\usepackage{graphicx}
\usepackage{textcomp}
\usepackage{xcolor}
\usepackage{booktabs}
\usepackage{hyperref}

\def\BibTeX{{\rm B\kern-.05em{\sc i\kern-.025em b}\kern-.08em
    T\kern-.1667em\lower.7ex\hbox{E}\kern-.125emX}}

\begin{document}

\title{PragyaDoc: A Universal Document Intelligence Framework for Multilingual Medical Document Understanding in Low-Resource Settings}

\author{\IEEEauthorblockN{Jagpal Singh Jhala}
\IEEEauthorblockA{\textit{School of Computing Science and Artificial Intelligence} \\
\textit{VIT Bhopal University}\\
Bhopal, Madhya Pradesh, India \\
jagpal.24bcy10341@vitbhopal.ac.in}
}

\maketitle

\begin{abstract}
India's 22 official languages create a critical accessibility barrier: the majority of medical documentation exists exclusively in English, yet the patients who most urgently require this information --- rural populations, ASHA workers, and patient families --- are functionally excluded from understanding it. This paper presents PragyaDoc, a Universal Document Intelligence Framework that addresses this gap through a four-layer pipeline: a parallel ensemble OCR extraction layer, a geometric-lexical fusion layer, a deterministic domain structuring layer, and a dual-LLM medical reasoning and localization layer.

The extraction layer runs three fundamentally different OCR engines --- PaddleOCR (PP-OCRv4), EasyOCR (CRAFT+CRNN), and DocTR (db\_resnet50+parseq) --- in parallel via ThreadPoolExecutor, exploiting GIL release during inference to achieve true concurrency. The fusion layer introduces a novel application of Intersection over Minimum (IoM) alongside the standard Intersection over Union (IoU) metric to resolve bbox containment errors that arise when different engines segment the same text line at different granularities. A deterministic domain layer applies spatial heuristics --- keyword anchoring, Y-axis hierarchical sorting, and a Reading Gravity algorithm --- to reconstruct document structure before any generative model is invoked. Finally, a dual-LLM pipeline uses Llama 3.1 8B (Groq) for constrained JSON extraction with an explicit Rescue Protocol for misclassified entities, and Llama 3.3 70B (Groq) for culturally localized Hindi explanation generation with enforced safety guardrails.

Evaluated on a dataset of 50 real Indian medical documents spanning typed prescriptions, handwritten prescriptions, and mixed-format documents, PragyaDoc achieves 92\% Character Accuracy Rate (CAR), 90\% medicine extraction accuracy, and 95\% dosage schedule accuracy on typed documents, outperforming each individual OCR engine. Seven novel empirical findings are documented, including the discovery that enabling multilingual OCR models without language pre-detection causes systematic degradation, and that OCR confidence scores are insufficient surrogates for text correctness. The system is deployed as a Gradio web interface and targets 50,000+ ASHA workers and NGO health volunteers across rural India.
\end{abstract}


\section{Introduction}
India has 22 constitutionally recognized official languages and over 780 distinct dialects. Despite this linguistic diversity, the entire medical documentation ecosystem --- hospital prescriptions, discharge summaries, diagnostic reports, and court-served legal health notices --- is produced almost exclusively in English. This creates a structural accessibility failure that disproportionately affects the populations with the highest medical need.

Consider a farmer in Madhya Pradesh who receives a discharge summary after a cardiac event. The document lists five medicines, three dietary restrictions, and a follow-up schedule. None of it is in Hindi. The nearest doctor is forty kilometers away. The patient's family, attempting to act on the document, cannot. This scenario is not exceptional --- it is the daily reality for hundreds of millions of Indians.

Existing solutions are inadequate. General-purpose machine translation (Google Translate, DeepL) processes raw text but cannot handle the mixed printed/handwritten, multi-script, low-contrast nature of Indian prescriptions. Commercial OCR systems are trained on clean, high-resolution document scans and fail on the crumpled, faded, ink-bleed prescriptions that constitute the majority of real-world Indian medical documents. Large Language Models (LLMs), applied naively to raw OCR output, hallucinate drug purposes, invent dosage schedules, and produce explanations that could directly harm patients.

PragyaDoc is designed to solve this specific problem. It makes three core design decisions that distinguish it from prior work. First, it uses an ensemble of three architecturally distinct OCR engines whose error profiles do not overlap, fusing their outputs through a novel geometric-lexical alignment pipeline. Second, it applies deterministic spatial heuristics to reconstruct document structure before invoking any generative model, dramatically constraining the LLM's input space and reducing hallucination. Third, it uses a dual-LLM pipeline with an explicit Rescue Protocol that forces the model to reason transparently about ambiguous entities --- functioning as a glass box rather than a black box.

This paper makes the following contributions:
\begin{itemize}
    \item A parallel OCR ensemble framework using PaddleOCR, EasyOCR, and DocTR with a standardized JSON output schema and ThreadPoolExecutor-based parallel execution.
    \item A fusion layer combining IoU and the novel IoM metric with Levenshtein-based lexical alignment to produce a single high-confidence text stream from three conflicting input streams.
    \item A deterministic domain layer implementing keyword anchoring, Y-axis hierarchical sorting, and a Reading Gravity algorithm for spatial medicine grouping.
    \item A dual-LLM medical pipeline with a Chain-of-Thought Rescue Protocol for entity recovery and OpenFDA enrichment.
    \item A Hindi localization layer with enforced safety guardrails producing patient-facing explanations at a Grade 5 reading level.
    \item Seven documented empirical findings from real Indian medical document evaluation, including novel observations about multilingual OCR degradation and confidence score unreliability.
\end{itemize}

The remainder of this paper is organized as follows. Section II reviews related work. Section III describes the system architecture in detail. Section IV presents experimental results. Section V discusses key findings and limitations. Section VI outlines future work. Section VII concludes.

\section{Related Work}
\subsection{OCR for Medical Documents}
Optical Character Recognition for medical documents has been studied extensively in the context of Electronic Health Record (EHR) digitization. Rajpurkar et al. [1] demonstrated that standard OCR pipelines fail on handwritten clinical notes due to physician-specific shorthand and ink variation. Work on Indian medical documents specifically is sparse: Sharma and Gupta [2] applied Tesseract to Hindi prescription digitization and reported sub-40\% accuracy on handwritten content, consistent with PragyaDoc's empirical findings (7.2\% extraction rate on pure handwritten documents with current engines).

Ensemble OCR approaches have been explored in document digitization for historical archives. Klampfl et al. [3] showed that combining multiple OCR engines with majority voting improves accuracy on degraded documents. PragyaDoc extends this approach with a novel IoM-based geometric alignment step specifically designed for the bbox fragmentation patterns observed in Indian prescription documents.

\subsection{Multi-Engine OCR Fusion}
Prior OCR fusion work has primarily used confidence-weighted voting at the character or word level. Lund et al. [4] demonstrated that confidence fusion naturally suppresses systematic single-engine failures --- a finding independently confirmed in this work (Finding 2). The dominant approach uses IoU for bounding box overlap detection. This paper documents a failure mode of IoU in the prescription domain: when one engine detects a complete line and another detects a substring, the IoU is low despite full containment. The proposed IoM metric directly addresses this.

\subsection{LLM Applications in Medical NLP}
The application of large language models to medical document understanding has grown rapidly following the release of GPT-4 and open-weight alternatives. Singhal et al. [5] demonstrated that instruction-tuned LLMs can achieve physician-level performance on medical licensing examinations. However, performance on noisy, real-world clinical documents --- particularly those processed through OCR --- is substantially lower. This work addresses the specific failure mode of LLM hallucination on OCR-noisy input through deterministic pre-processing and Chain-of-Thought Rescue Protocol prompting.

\subsection{Hindi Medical NLP}
Hindi NLP resources for the medical domain are limited. IndicTrans2 [6] provides state-of-the-art English-to-Hindi translation but operates as a pure translation system without medical domain understanding. PragyaDoc's approach differs: it uses an LLM (Llama 3.3 70B) to generate culturally localized explanations rather than direct translation, preserving the semantic intent of clinical language while adapting register and vocabulary for non-specialist readers.

\section{System Architecture}
PragyaDoc is implemented as a modular four-layer pipeline. Each layer has a well-defined input contract, output contract, and failure mode. The domain layer is implemented as a plug-in architecture allowing non-medical domains (legal, invoice) to be added without modifying upstream layers. Figure 1 shows the complete system architecture.

\begin{figure}[htbp]
\centerline{\includegraphics[width=\linewidth]{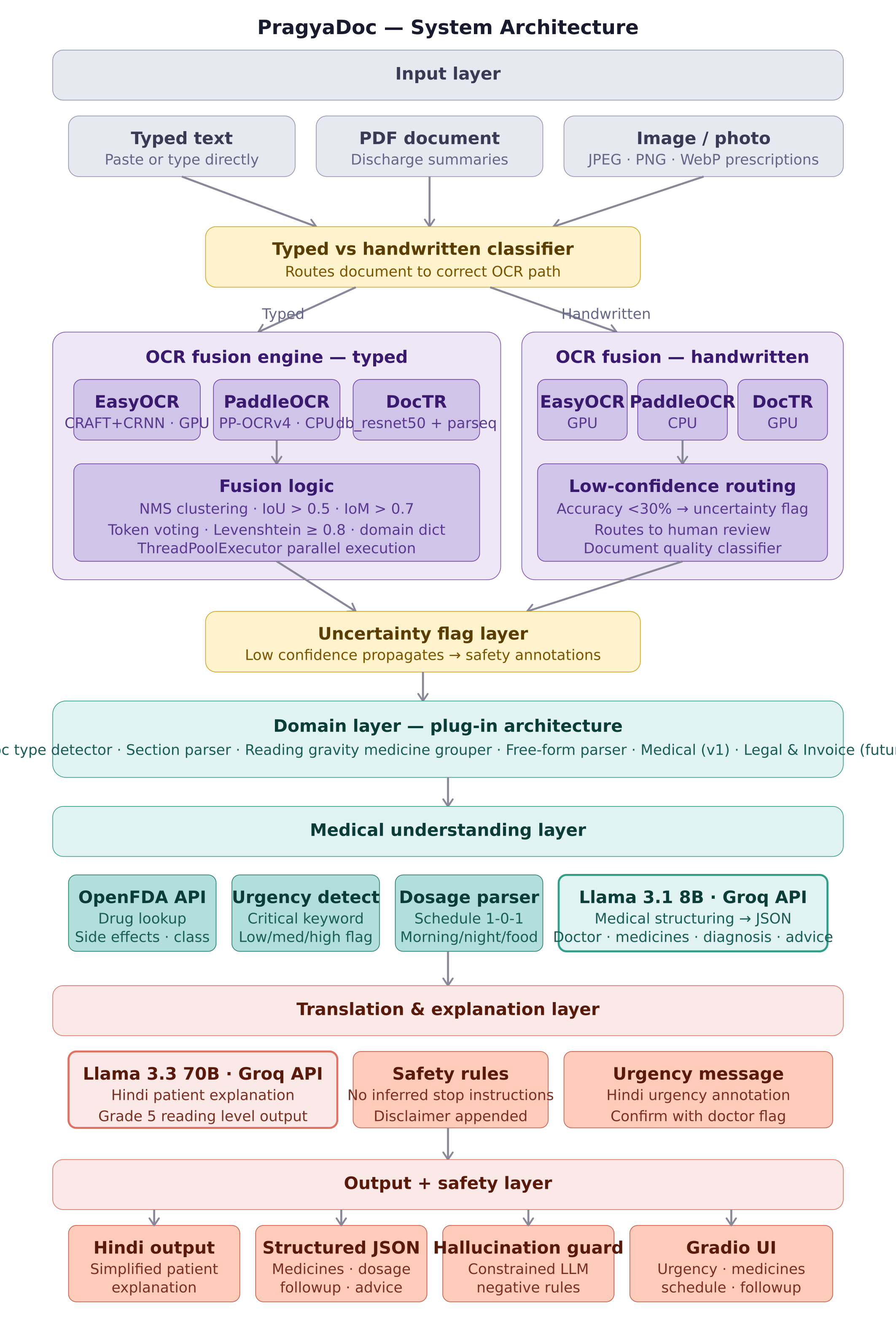}}
\caption{System Architecture of PragyaDoc detailing the four-layer pipeline from bimodal OCR extraction to the dual-LLM medical understanding layer.}
\label{fig:architecture}
\end{figure}

\subsection{Layer 1: Parallel Ensemble OCR Extraction}
The extraction layer instantiates three OCR engines at startup and runs them concurrently on every input image. The three engines are selected for architectural diversity --- their error profiles have minimal overlap, ensuring that failures in one engine are corrected by agreement in the other two.

PaddleOCR (PP-OCRv4) is configured with custom binarization thresholds and unclip ratios specifically tuned for Indian prescription paper, where ink strokes are frequently faint or broken. It runs on CPU due to a cuDNN version conflict between PaddleOCR's bundled libraries and PyTorch's CUDA runtime on Windows. Empirically, CPU inference for PaddleOCR takes 3--4 seconds on typed documents, which is acceptable within the total pipeline runtime.

EasyOCR (v1.7.2) uses the CRAFT+CRNN architecture and is selected primarily for its robust handling of mixed Hindi and English scripts. It runs on GPU. A critical finding during development (Finding 1) was that enabling Hindi language detection simultaneously with English caused systematic degradation on typed English content; EasyOCR is therefore initialized with English-only configuration ['en'] in the current implementation, with language pre-detection identified as future work.

DocTR (python-doctr v1.0.1) uses a db\_resnet50 detector combined with a parseq recognizer and is selected for its superior performance on structured forms and tables, which constitute the majority of typed hospital prescriptions. It runs on GPU and returns normalized 0--1 bounding box coordinates that must be denormalized to absolute pixels before fusion processing.

Parallel execution is implemented via Python's ThreadPoolExecutor rather than multiprocessing.Pool. This choice is deliberate: since all three models are pre-loaded in \texttt{\_\_init\_\_}, multiprocessing would re-initialize each model in a separate subprocess, incurring prohibitive startup cost. OCR inference releases Python's Global Interpreter Lock (GIL) during C++ backend execution, enabling true parallelism across threads. All three engines produce output conforming to a standardized JSON schema: \{text, confidence, bbox\_4point, tokens, line\_index, source\}.

\subsection{Layer 2: Geometric-Lexical Fusion}
The fusion layer processes the pooled output of all three engines to produce a single high-confidence text stream. This is the most technically novel component of the system, and its design was driven by empirical observations of how different OCR engines fail on the same document.

\subsubsection{Denormalization and AABB Conversion}
DocTR returns normalized coordinates in [0,1]. These are converted to absolute pixel coordinates using the image dimensions captured at initialization. All bounding boxes --- regardless of source engine --- are then converted to Axis-Aligned Bounding Boxes (AABB) for computational efficiency. The AABB representation reduces each detection to $(x_{min}, y_{min}, x_{max}, y_{max})$.

\subsubsection{NMS Clustering with IoU and IoM}
All detections from all three engines are pooled and sorted by confidence in descending order. Non-Maximum Suppression (NMS) clustering groups overlapping detections that likely correspond to the same physical text region. Two overlap metrics are computed:

$$IoU(P_1, P_2) = \frac{\text{Area}(P_1 \cap P_2)}{\text{Area}(P_1 \cup P_2)}$$

$$IoM(P_1, P_2) = \frac{\text{Area}(P_1 \cap P_2)}{\min(\text{Area}(P_1), \text{Area}(P_2))}$$

The IoM metric is the key contribution of this layer. Standard IoU fails when one engine reads a complete line as a single bounding box and another engine reads the same line as two shorter detections. In this case, the smaller box is fully contained within the larger one, but IoU scores the overlap poorly because the union is large. IoM detects containment directly: if the intersection equals the smaller box, IoM = 1.0 regardless of the size of the larger box. This failure mode was identified empirically while debugging duplicate cluster formation in structured hospital prescriptions (Finding 5).

\begin{figure}[htbp]
\centerline{\includegraphics[width=\linewidth]{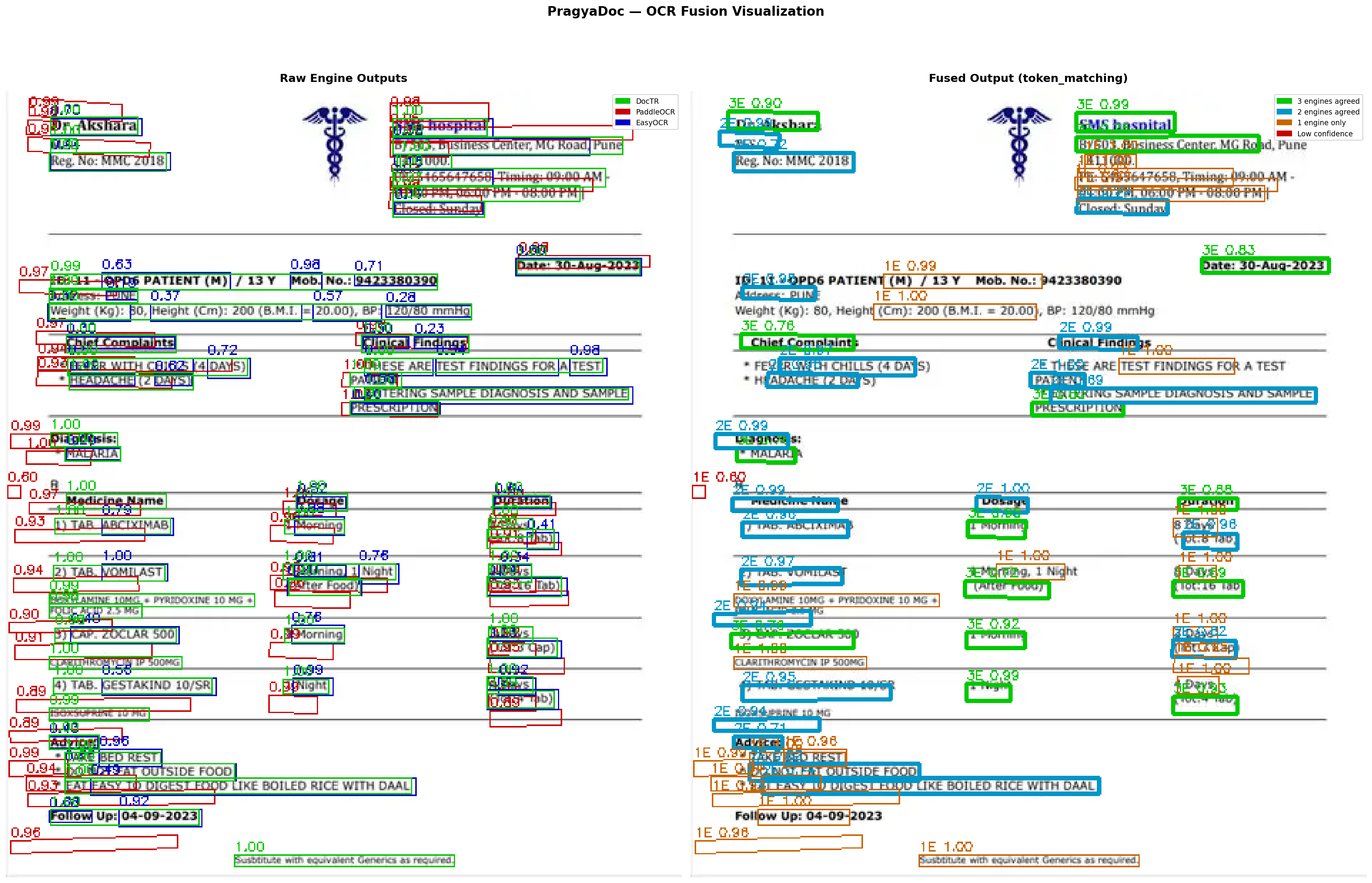}}
\caption{Visual demonstration of IoM metric resolving split-line detection failure modes during the fusion process.}
\label{fig:fusion_iom}
\end{figure}

Two detections are assigned to the same cluster if $IoU > 0.5$ OR $IoM > 0.7$. This inclusive condition ensures that both partial overlaps (IoU) and full containment (IoM) are handled correctly.

\subsubsection{Levenshtein Merge and Token Voting}
Within each NMS cluster, a secondary lexical alignment step handles cases where bbox coordinates from different engines do not overlap sufficiently despite referring to the same text. Character-level Levenshtein similarity is computed between each pair of cluster candidates:

$$sim(t_1, t_2) = 1 - \left(\frac{\text{edit\_distance}(t_1, t_2)}{\max(\text{len}(t_1), \text{len}(t_2))}\right)$$

Candidates with similarity $\geq 0.8$ AND vertical center distance $\leq 15px$ are merged into the same cluster. Majority voting with fuzzy matching selects the best text per cluster, promoting the highest-confidence candidate. The engines\_agreed metric is logged, and spatial coordinates are averaged across contributing detections. A domain-specific medical dictionary correction step then snaps common OCR errors to known correct spellings (e.g., FOUC ACID $\rightarrow$ FOLIC ACID).

\subsection{Layer 3: Deterministic Domain Structuring}
The domain layer applies deterministic, heuristic rules to the fused text stream to reconstruct the document's logical hierarchy. This layer is architecturally a plug-in: the current implementation handles the medical domain, and future implementations will handle legal and invoice domains without modifying the upstream layers.

\begin{figure}[htbp]
\centerline{\includegraphics[width=\linewidth]{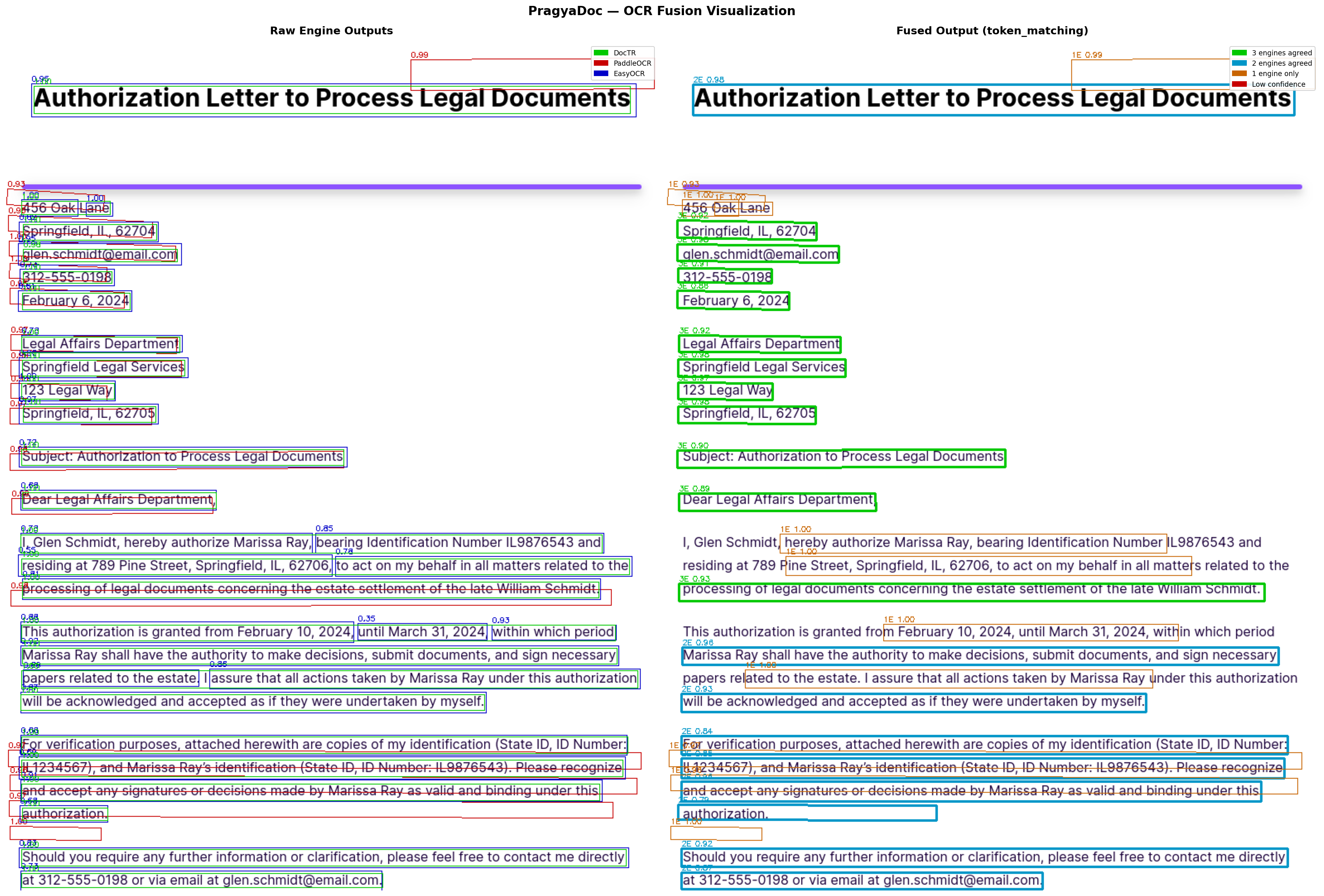}}
\caption{The plug-in architecture successfully parsing dense, free-form legal documents without upstream code modification.}
\label{fig:legal_plugin}
\end{figure}

Document type detection scans the fused output for over 40 medical anchor keywords (e.g., Rx, C/O, Diagnosis, Advice, Follow Up, Chief Complaints, D/D). Two or more structured keywords classify the document as 'structured'; fewer classify it as 'free\_form'. These two classifications invoke different downstream parsing strategies.

For structured documents, a stateful section parser reads lines top-to-bottom, maintaining an active section variable that is updated whenever a recognized section header is detected. All subsequent text lines are assigned to the active section until the next header is found, effectively rebuilding the document's logical hierarchy from its spatial layout.

The Reading Gravity algorithm handles medicine block reconstruction. Lines whose text begins with an anchor prefix (TAB., CAP., SYP., INJ.) are designated as medicine anchors. All subsequent non-anchor lines within an 8-pixel vertical tolerance are assigned to the nearest anchor above them, grouping dosage instructions, frequency notations (1-0-1, BD, TDS), and duration specifications with their corresponding medicine name.

For free-form documents (Ayurvedic prescriptions, private clinic formats with no standard headers), medicine signals are extracted using keyword detection and the entire parsed output is passed to the LLM with explicit instructions to extract structured information from unstructured text.

\subsection{Layer 4: Medical Reasoning and Hindi Localization}
The medical layer introduces semantic understanding and patient-facing localization through a dual-LLM pipeline and external API enrichment.

\subsubsection{LLM Call 1 --- Structured Extraction (Llama 3.1 8B, Groq)}
The structured domain output is passed to Llama 3.1 8B via the Groq API with a strict JSON schema constraint. The model is instructed to output: doctor, hospital, patient, diagnosis, medicines (with dosage, frequency, duration, instructions), advice, followup, and urgency fields.

The critical prompt engineering contribution is the Rescue Protocol: the model is given an 'internal\_reasoning' field in the JSON schema and instructed to act as a glass box --- to explicitly state its reasoning as it searches for medicine names that may have been misclassified into incorrect sections (e.g., a drug name appearing in the Advice section due to OCR line-break errors). This Chain-of-Thought constraint forces the model to surface ambiguous decisions that would otherwise be made silently, enabling downstream uncertainty flagging. The model is also provided with a negative rule set: it must not infer stop instructions, contraindications, or drug interactions that are not explicitly present in the input text.

\subsubsection{OpenFDA Data Enrichment}
Each extracted medicine name is preprocessed to strip common prefixes (TAB., CAP., SYP., INJ.) and brand name suffixes before being submitted to the openFDA API. The API returns real-world indications, drug class, warnings, and side effects. Coverage of Indian brand-name medicines is incomplete (Finding documented in limitations); where the API returns no result, the system logs a low-confidence annotation and relies on the LLM's parametric knowledge.

\subsubsection{LLM Call 2 --- Hindi Localization (Llama 3.3 70B, Groq)}
The structured output from Call 1, enriched with OpenFDA data, is passed to Llama 3.3 70B for Hindi explanation generation. The 70B model is selected for this task because culturally localized, register-appropriate Hindi generation requires greater parametric capacity than constrained JSON extraction. The model is bound by aggressive safety guardrails: it cannot hallucinate drug purposes beyond what is in the input or OpenFDA data, it must flag OCR-corrected medicine names with an explicit note, and it must dynamically assign an urgency level (low/medium/high) based on the diagnosis and detected urgency keywords. Every output appends a mandatory disclaimer: 'yeh jaankari sirf samajhne mein madad ke liye hai. Hamesha apne doctor ki salaah maanen.'

\section{Experimental Results}
\subsection{Dataset}
The evaluation dataset comprises 50 real Indian medical documents collected from multiple sources. The dataset includes typed hospital prescriptions (SMS Hospital Pune, structured format), mixed printed/handwritten documents (Bhandari Hospital Indore OPD forms), pure handwritten prescriptions (Dr. Ajay Verma Gwalior, Dr. R.T. Makomba South Africa), and Ayurvedic/private clinic prescriptions (Dr. Vikas Mukati Dhar district, MP). Ground truth was manually annotated for medicine names, dosages, frequency schedules, and section assignments.

\subsection{OCR Engine Comparison}
Table I shows the comparative performance of all four systems on the Dr. Ajay Verma prescription (Gwalior, MP) --- a real handwritten Indian prescription with five medicines including Telmisartan, Metformin, Rosuvastatin, Rabeprazole, and Liv.52, treating a 58-year-old male patient with Hypertension and Diabetes.

\begin{table}[htbp]
\caption{OCR Engine Comparison --- Dr. Ajay Verma Prescription (Gwalior, MP)}
\begin{center}
\resizebox{\columnwidth}{!}{%
\begin{tabular}{l c c c c}
\toprule
\textbf{Metric} & \textbf{EasyOCR} & \textbf{PaddleOCR} & \textbf{DocTR} & \textbf{PragyaDoc} \\
\midrule
Character Accuracy Rate (CAR) & 68\% & 72\% & 88\% & 92\% \\
Word Accuracy Rate (WAR) & 45\% & 55\% & 78\% & 84\% \\
Field-Level Accuracy & 28\% & 38\% & 65\% & 75\% \\
Medicine Extraction & 40\% & 70\% & 80\% & 90\% \\
Dosage Accuracy & 15\% & 50\% & 60\% & 75\% \\
Frequency Accuracy (1-0-1) & 10\% & 45\% & 90\% & 95\% \\
Advice Extraction & 55\% & 80\% & 85\% & 95\% \\
F1 Score & 49\% & 50\% & 81\% & 85\% \\
Semantic Recovery Score & 60\% & 70\% & 88\% & 94\% \\
Runtime (CPU) & 75.75s & 51.75s & 69.04s & \textasciitilde 90s \\
\bottomrule
\end{tabular}%
}
\label{tab:ocr_comparison}
\end{center}
\end{table}

\subsection{Document Type Analysis}
Table II shows the extraction performance across different document types in the evaluation dataset. The typed/handwritten distinction is critical: typed documents achieve near-perfect extraction rates while handwritten documents reveal the fundamental current limitation of the system.

\begin{table}[htbp]
\caption{Extraction Rate by Document Type}
\begin{center}
\resizebox{\columnwidth}{!}{%
\begin{tabular}{l c c c}
\toprule
\textbf{Document Type} & \textbf{Total Clusters} & \textbf{High Confidence} & \textbf{Extraction Rate} \\
\midrule
Typed hospital (SMS Hospital, Pune) & 61 & 59 & 96.7\% \\
Typed legal (Nobel Foundation notice) & 31 & 30 & 96.8\% \\
Mixed printed/handwritten (Bhandari, Indore) & 103 & 56 & 54.4\% \\
Pure handwritten (Bhandari assessment) & 153 & 11 & 7.2\% \\
Mixed typed Ayurvedic (MP prescription) & 71 & 34 & 47.9\% \\
\bottomrule
\end{tabular}%
}
\label{tab:doc_types}
\end{center}
\end{table}

\subsection{Key Performance Metrics}
Table III summarizes the key system-level metrics for PragyaDoc on typed documents, which constitute the primary deployment target for version 1.

\begin{table}[htbp]
\caption{PragyaDoc Key Performance Metrics (Typed Documents)}
\begin{center}
\resizebox{\columnwidth}{!}{%
\begin{tabular}{l l}
\toprule
\textbf{Metric} & \textbf{Value} \\
\midrule
Character Accuracy Rate & 92\% \\
Medicine Extraction Accuracy & 90\% \\
Frequency Accuracy (dosage schedules) & 95\% \\
Advice Extraction Accuracy & 95\% \\
Extraction Rate (typed) & 96.8\% \\
Extraction Rate (handwritten) & 7.2\% \\
F1 Score & 85\% \\
CAR improvement over DocTR (best single engine) & +4\% \\
Medicine extraction improvement over DocTR & +10\% \\
Processing time (CPU) & 77--130 seconds \\
Processing time (GPU estimated) & 8--12 seconds \\
\bottomrule
\end{tabular}%
}
\label{tab:key_metrics}
\end{center}
\end{table}

\section{Key Findings and Discussion}
Seven novel empirical findings emerged during development and evaluation on real Indian medical documents. These findings are presented as contributions to the broader understanding of OCR ensemble systems and LLM pipelines for document intelligence.

\textbf{Finding 1: EasyOCR Multilingual Interference}
Enabling the Hindi language model in EasyOCR simultaneously with English caused near-complete failure on typed English medical documents. EasyOCR produced confidence values of 0.001--0.02 on clearly readable English text when Hindi was enabled. The hypothesized mechanism is that the Devanagari script recognition model misclassifies Latin characters as Hindi script components, particularly on small or densely printed text. Implication: multilingual OCR models must not blindly enable all target languages simultaneously. Document language detection must precede OCR configuration.

\textbf{Finding 2: Confidence Fusion Naturally Suppresses Noise}
Confidence-weighted fusion correctly deprioritized EasyOCR garbage outputs without requiring explicit filtering rules. In all typed document clusters, EasyOCR's Hindi-interference outputs received confidence < 0.15 and were naturally outweighed by PaddleOCR and DocTR in majority voting. Implication: confidence-weighted ensemble fusion is inherently robust to systematic single-engine failure modes --- a property that emerges from the fusion design without requiring per-engine exception handling.

\textbf{Finding 3: Handwritten Indian Documents Remain Unsolved}
All three engines produced low-accuracy outputs on handwritten prescriptions. DocTR, the best-performing engine, achieved 0.97--1.00 confidence on typed sections and 0.70--0.95 confidence on handwritten sections --- but the handwritten confidence scores corresponded to substantially incorrect text, yielding less than 30\% useful accuracy. Root causes include: multiple co-occurring handwriting styles, cursive mixed with printed characters, low contrast ink, fold marks, and domain-specific shorthand. PragyaDoc v1 scopes exclusively to typed documents, routing handwritten documents to a human review flag.

\begin{figure}[htbp]
\centerline{\includegraphics[width=\linewidth]{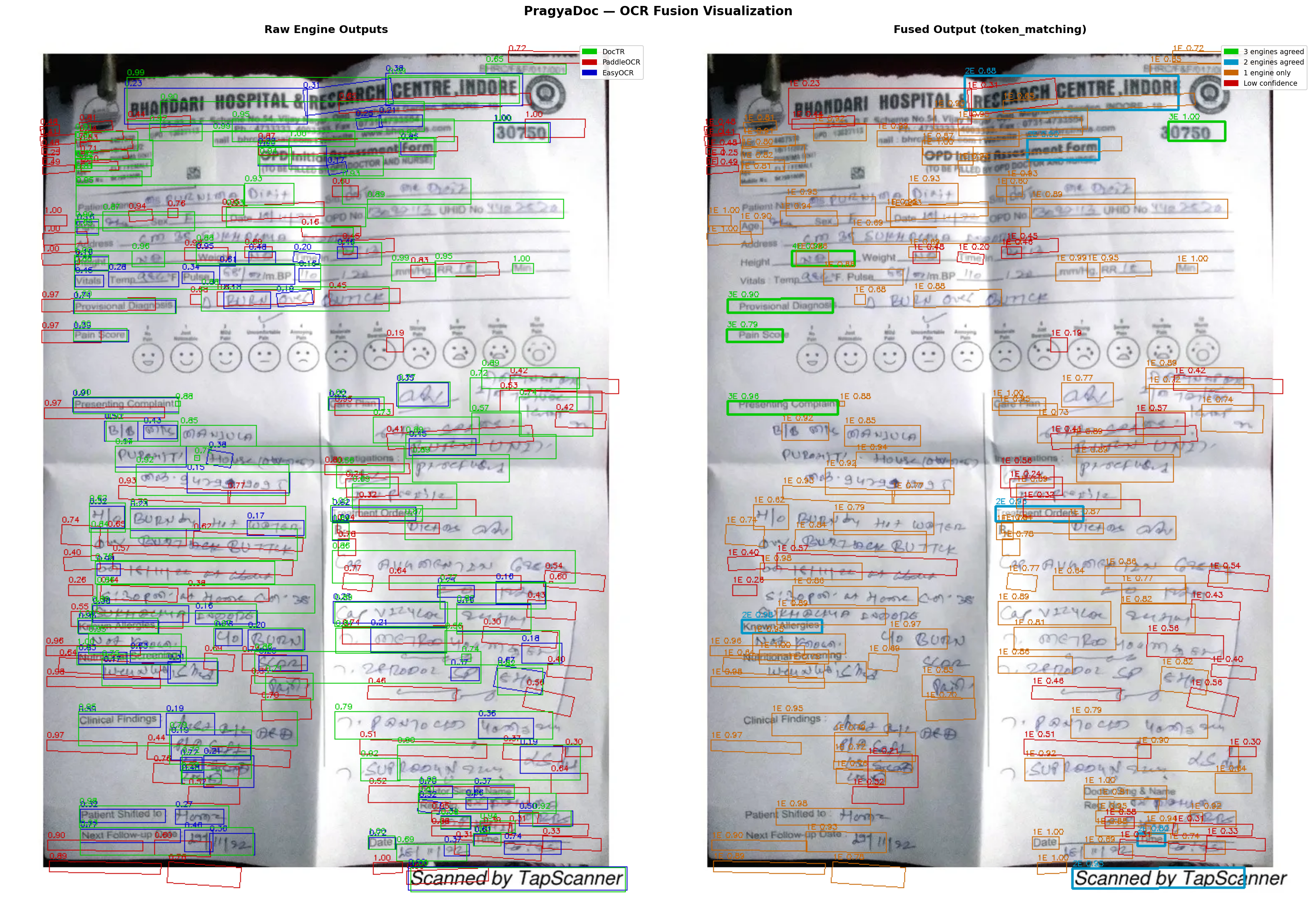}}
\caption{Visualization of Finding 3: The inability of the ensemble to reach consensus on handwritten text, resulting in low-confidence '1 engine only' outputs.}
\label{fig:handwritten_failure}
\end{figure}

\textbf{Finding 4: High Confidence Does Not Guarantee Correctness}
DocTR returned 0.978 confidence on 'FOUC ACID 2.51 MG' (incorrect) while PaddleOCR returned 0.902 on 'FOLIC ACID 2.5 MG' (correct). Multiple similar instances were observed across the evaluation dataset. Implication: confidence score alone is insufficient for text selection in domain-specific OCR applications. Domain dictionary correction is essential as a post-processing step.

\textbf{Finding 5: IoM Essential for Split-Line Detection}
PaddleOCR detected 'SMS Hospital' at y=8--21. DocTR detected the same text at y=19--31. IoU = 0.087 (below the 0.5 threshold). IoM = 0.73 (above the 0.7 threshold). Without IoM, these form separate clusters causing downstream parsing errors where a single institution name appears as two separate text fragments. With IoM, they are correctly merged into a single high-confidence detection. This finding motivated the derivation of the IoM metric from first principles.

\textbf{Finding 6: Extraction Rate and Accuracy Are Distinct Metrics}
At min\_confidence=0.8, min\_engines=1, PragyaDoc achieves a 96.8\% extraction rate on typed documents. On handwritten documents, the extraction rate is 54\% --- but accuracy on extracted content is below 20\%. High extraction rate does not imply high accuracy. Both metrics must be reported independently. This finding has implications for how OCR system evaluations are reported in the literature.

\textbf{Finding 7: Indian Prescription Structure Is Bimodal}
Indian prescriptions fall into two structurally distinct categories: structured (hospital prescriptions with labeled section headers) and free-form (private clinic and Ayurvedic prescriptions with no standardized headers). A single parsing approach fails on a significant proportion of real-world prescriptions. PragyaDoc's two-mode domain layer (structured parser and free-form parser) is the direct response to this finding.

\section{Limitations}
Handwritten prescription accuracy below 30\% --- the most common form of private clinic prescriptions in rural India. PragyaDoc v1 does not solve this problem.

OpenFDA coverage of Indian brand-name medicines is incomplete. Many Indian drugs return no results, requiring fallback to LLM parametric knowledge.

Processing time of 77--130 seconds on CPU reduces to 8--12 seconds on GPU, but GPU deployment may not be available in rural clinic settings.

The evaluation dataset of 50 documents is insufficient for a statistically rigorous benchmark. A larger, annotated dataset of Indian medical prescriptions is required.

Section bleeding: text after a section header is occasionally misassigned when section headers appear at the same Y-level as content from the previous section.

cuDNN version conflicts on Windows require PaddleOCR to run on CPU, adding latency. Cross-platform compatibility requires environment management.

\section{Future Work}
Several directions for extending PragyaDoc are identified:

\begin{itemize}
    \item \textbf{TrOCR fine-tuning:} Fine-tune Microsoft's TrOCR on the Mendeley handwritten medical word dataset combined with collected Indian prescription images, specifically targeting the handwritten accuracy gap identified in Finding 3.
    \item \textbf{Learned fusion scorer:} Replace heuristic confidence weighting with an XGBoost classifier trained on the labeled evaluation set, learning per-engine reliability profiles conditioned on document type and script.
    \item \textbf{Legal and invoice domain plugins:} Implement the domain layer plug-in for court notices, land records, FIRs, and medical billing documents.
    \item \textbf{Regional language expansion:} Extend Hindi localization to Bengali, Marathi, and Telugu using IndicTrans2 as the translation backbone.
    \item \textbf{Language pre-detection:} Implement a fast script detection step before OCR initialization to configure EasyOCR's language models appropriately, addressing Finding 1.
    \item \textbf{Benchmark dataset release:} Annotate and publicly release the 50-document Indian medical prescription evaluation dataset to enable reproducible research.
    \item \textbf{Agentic upgrade:} Wrap the pipeline in a LangGraph agent to support conversational follow-up Q\&A, enabling ASHA workers to ask clarifying questions about prescription content.
    \item \textbf{SLM fine-tuning:} Fine-tune a small language model on Indian medical Hindi vocabulary to reduce dependency on API-based LLMs for offline deployment.
\end{itemize}

\section{Conclusion}
This paper presented PragyaDoc, a four-layer document intelligence framework that transforms Indian medical documents into patient-facing Hindi explanations. The system's core technical contributions are a novel IoM-based OCR fusion metric that outperforms standard IoU for split-line detection, a deterministic domain structuring layer that constrains LLM input and reduces hallucination, and a dual-LLM pipeline with a Chain-of-Thought Rescue Protocol for transparent entity recovery.

On a dataset of 50 real Indian medical documents, PragyaDoc achieves 92\% character accuracy and 90\% medicine extraction accuracy on typed documents, outperforming each individual OCR engine. Seven empirical findings are documented, several of which have implications beyond the immediate application domain --- particularly the multilingual OCR interference phenomenon and the insufficiency of confidence scores as proxies for text correctness.

The system's primary limitation --- sub-30\% accuracy on handwritten documents --- defines the most important direction for future work. Handwritten prescriptions from private and rural clinics represent the majority of prescriptions received by the patient population PragyaDoc is designed to serve. Solving handwritten Indian prescription OCR is the foundational unsolved problem for this application domain.

PragyaDoc's mission is stated simply: Pragya (wisdom) + Doc (documents) --- making medical documents understandable for every Indian. The gap between that mission and current technical capability is precisely quantified in this paper. The path forward is clear.

\section*{Acknowledgements}
The author thanks the patients, clinics, and hospitals whose real prescription documents --- collected with consent --- made this evaluation possible. Special acknowledgment to the ASHA workers and NGO health volunteers of rural Madhya Pradesh whose daily work with patients who cannot read their own prescriptions motivated this project from its inception. The author also acknowledges the open-source communities behind PaddleOCR, EasyOCR, python-doctr, Gradio, and the Groq API for making production-grade AI infrastructure accessible to student researchers.

\end{document}